\documentclass[11pt,letterpaper]{article}
\usepackage{emnlp2017}
\usepackage{times}
\usepackage{latexsym}

\usepackage{url}

\usepackage{amsmath,graphicx,bm} 
\usepackage{array,multirow}
\usepackage{algorithm}
\usepackage{algorithmic}
\usepackage{bbold}
\usepackage{subfigure}

\usepackage{amssymb}
\usepackage{xcolor}
\usepackage{colortbl}

\emnlpfinalcopy

\def\emnlppaperid{815}

\DeclareMathOperator*{\argmax}{arg\,max}
\newcolumntype{P}[1]{>{\centering\arraybackslash}p{#1}}

\title{Reinforcement Learning with External Knowledge and Two-Stage Q-functions for Predicting Popular Reddit Threads}

\author{Ji He, Mari Ostendorf \\
Department of Electrical Engineering\\
University of Washington\\
Seattle, WA 98195, USA \\
\texttt{\{jvking, ostendor\}@uw.edu}
\And
Xiaodong He \\
Microsoft Research \\
Redmond, WA 98052, USA \\
\texttt{\{xiaohe\}@microsoft.com}
}

\date{}

\begin{document}

\maketitle

\begin{abstract}
This paper addresses the problem of predicting popularity of comments in an online discussion forum using reinforcement learning, particularly addressing two challenges that arise from having natural language state and action spaces. First, the state representation, which characterizes the history of comments tracked in a discussion at a particular point, is augmented to incorporate the global context represented by discussions on world events available in an external knowledge source. Second, a two-stage Q-learning framework is introduced, making it feasible to search the combinatorial action space while also accounting for redundancy among sub-actions. We experiment with five Reddit communities, showing that the two methods improve over previous reported results on this task.
\end{abstract}

\section{Introduction}
\label{sec:intro}
Reinforcement learning refers to learning strategies for sequential decision-making tasks, where a system takes actions at a particular state with the goal of maximizing a long-term reward. Recently, several tasks that involve states and actions described by natural language have been studied, such as text-based games \cite{narasimhan-kulkarni-barzilay:2015:EMNLP,he2016acl}, web navigation \cite{nogueira2016webnav}, information extraction \cite{narasimhan-yala-barzilay:2016:EMNLP2016}, Reddit popularity prediction and tracking \cite{he2016emnlp}, and human-computer dialogue systems \cite{wen2016network,li-EtAl:2016:EMNLP20162}. Some of these studies ignore the use of external knowledge or world knowledge, while others (such as information extraction and task-oriented dialogue systems) directly interact with an (often) static database.

External knowledge -- both general and domain-specific -- has been shown to be useful in many natural language tasks, such as in question answering \cite{yang2003structured,katz2005external,lin2002web}, information extraction \cite{agichtein2000snowball,etzioni2011open,wu2010open}, computer games \cite{branavan2012learning}, and dialog systems \cite{ammicht1999knowledge,yan-EtAl:2016:P16-11}. However, in reinforcement learning, incorporating external knowledge is relatively rare, mainly due to the domain-specific nature of reinforcement learning tasks, e.g. Atari games \cite{mnih2015human} and the game of Go \cite{silver2016mastering}. Of particular interest in our work is external knowledge represented by unstructured text, such as news feeds, Wikipedia pages, search engine results, and manuals, as opposed to a structured knowledge base.

Our study is conducted on the task of Reddit popularity prediction proposed in He et al. \shortcite{he2016emnlp}, which is a sequential decision-making problem based on a large-scale real-world natural language data set. In this task, a specified number of discussion threads predicted to be popular are recommended, chosen from a fixed window of recent comments to track. The authors proposed a reinforcement learning solution in which the state is formed by the collection of comments in the threads being tracked, and actions correspond to selecting a subset of new comments to follow (sub-actions) from the set of recent contributions to the discussion. Since comments are potentially redundant (multiple respondents can have similar reactions), the study found that sub-actions were best evaluated in combination.  The computational complexity of the combinatorial action space was sidestepped by random sampling a fixed number of candidates from the full action space. A major drawback of random sampling in this application is that popular comments are rare and easily missed.


We make two main contributions in this paper. The first is a novel architecture for incorporating unstructured external knowledge into reinforcement learning. More specifically, information from the original state is used to query the knowledge source (here, an evolving collection of documents corresponding to other online discussions about world events), and the state representation is augmented by the outcome of the query. Thus, the agent can use both the local context (reinforcement learning environment) and the global context (e.g. recent discussions about world news) when making decisions. Second, we propose to use a two-stage Q-learning framework that makes it feasible to explore the full combinatorial natural language action space. A first Q-function is used to efficiently generate a list of sub-optimal candidate actions, and a second more sophisticated Q-function reranks the list to pick the best action.


\section{Task}
\label{sec:task}
On Reddit, users reply to posts and other comments in a threaded (tree-structured) discussion. Comments (and posts) are associated with a \emph{karma score}, which is a combination of positive and negative votes from registered users indicating popularity of the comment. In prior work \cite{he2016emnlp}, popularity prediction in Reddit discussions (comment recommendation) is proposed for studying reinforcement learning with a large scale natural language action space. 
At each time step $t$, the agent receives a string of text that describes the state $s_t$ and several strings of text that describe the potential actions $\{a_t^i\}\in\mathcal{A}_t$ (new comments to consider). The agent attempts to pick the best action for the purpose of maximizing the long-term reward.
%
%
In a real-time scenario, the final karma of a comment is not immediately available, so prediction of popularity is based on the text in the comment as well as the context of discussion history. It is common that a lower karma comment will eventually lead to more discussion and popular comments in the future. Thus it is natural to formulate this task as a reinforcement learning problem.

More specifically, the set of comments that are being tracked at time $t$ is denoted as $M_t$. The state, action, and immediate rewards are defined as follows:
\begin{itemize}
\item State: all previously tracked comments, as well as the post (root node of the tree), i.e. $s_t=\{ M_0, M_1, \cdots , M_t\}$
\item Action: an action is taken when a total of $N$ new comments $\mathcal{C}_t=\{ c_{t,1},c_{t,2},\cdots ,c_{t,N}\}$, 
 appear as nodes in the subtrees of $M_t$, and the agent picks a set of $K$ comments to be tracked in the next time step: $a_t=M_{t+1}=\{c_t^1,c_t^2,\cdots,c_t^K\}$, $c_t^i\in \mathcal{C}_t$ and $c_t^i\ne c_t^j$ if $i\ne j$
\item Reward: $r_{t+1}$ is the accumulated karma scores\footnote{The karma score is observed from an archived version of the discussion, not immediately shown at the time of the comment, so not available to a real-time system.} in comments in $M_{t+1}$
\end{itemize}
In this task, because an action corresponds to a set of comments (sub-actions) chosen from a larger set of candidates, the action space is combinatorial. $\mathcal{C}_t$ and $\mathcal{A}_t$ are also time-varying, reflecting the flow of the discussion in the paths chosen.

The standard Q-learning defines a function $Q (s, a)$ as the expected return starting from $s$ and taking the action $a$:
\begin{align*}
Q(s,a)=\mathbb{E}\left\{\sum_{l=0}^{+\infty}\gamma^l r_{t+1+l} | s_t=s, a_t=a \right\}
\end{align*}
where $\gamma\in [0, 1]$ denotes a discount factor. The Q-function associated with an optimal policy can be found by the Q-learning recursion \cite{watkins1992q}:
\begin{align*}
	Q(s_t, a_t) 
		\leftarrow &
			Q(s_t, a_t) 
			+ \eta_t \cdot \big( r_{t+1} + \\ \nonumber
		&	\gamma \cdot \max_{a'\in \mathcal{A}_{t+1}}{Q(s_{t+1}, a')}-Q(s_t, a_t) \big)
\end{align*}
where $\eta_t$ is the learning rate of the algorithm. In He et al. \shortcite{he2016emnlp}, two deep Q-learning architectures are proposed, both with separate networks for the state and action spaces yielding embeddings $h_s$ and $h_a^i$, respectively. Those embeddings are combined with a general interaction function $g(\cdot)$ to approximate the Q-values, $Q(s_t, a_t^i) = g\left(h_s, h_a^i\right)$, as in He et al. \shortcite{he2016acl}, where the approach of using separate networks for natural language state and action spaces is termed a Deep Reinforcement Relevance Network (DRRN). 

\begin{figure*}[t]
  \centerline{
  	\hfill
	\subfigure[DRRN-Sum]{
	\includegraphics[width=0.46\textwidth]{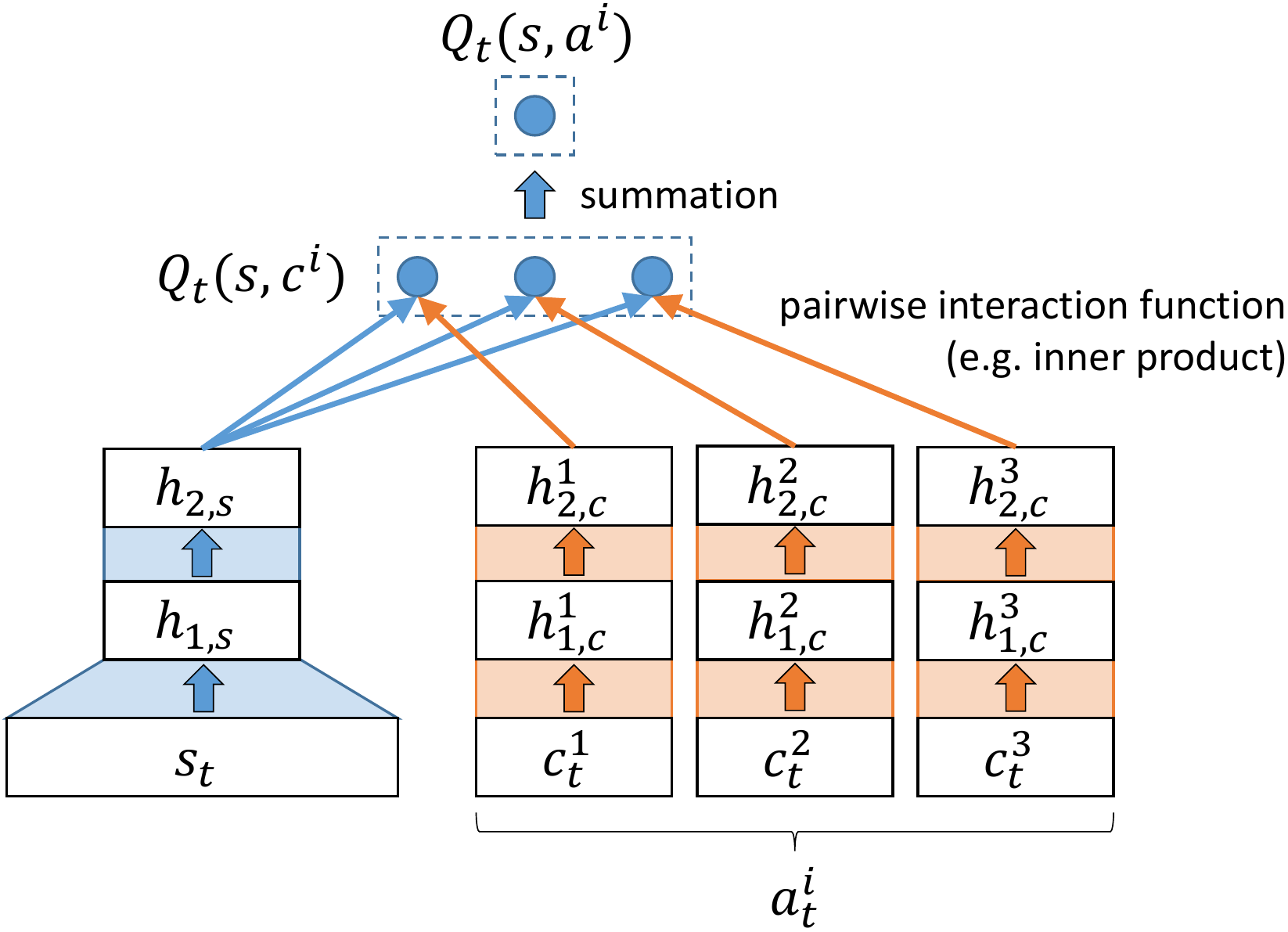}
  	\label{fig:architectures_drrn-sum}
  	}
  	\hfill
	\hfill
	\subfigure[DRRN-BiLSTM]{
	\includegraphics[width=0.39\textwidth]{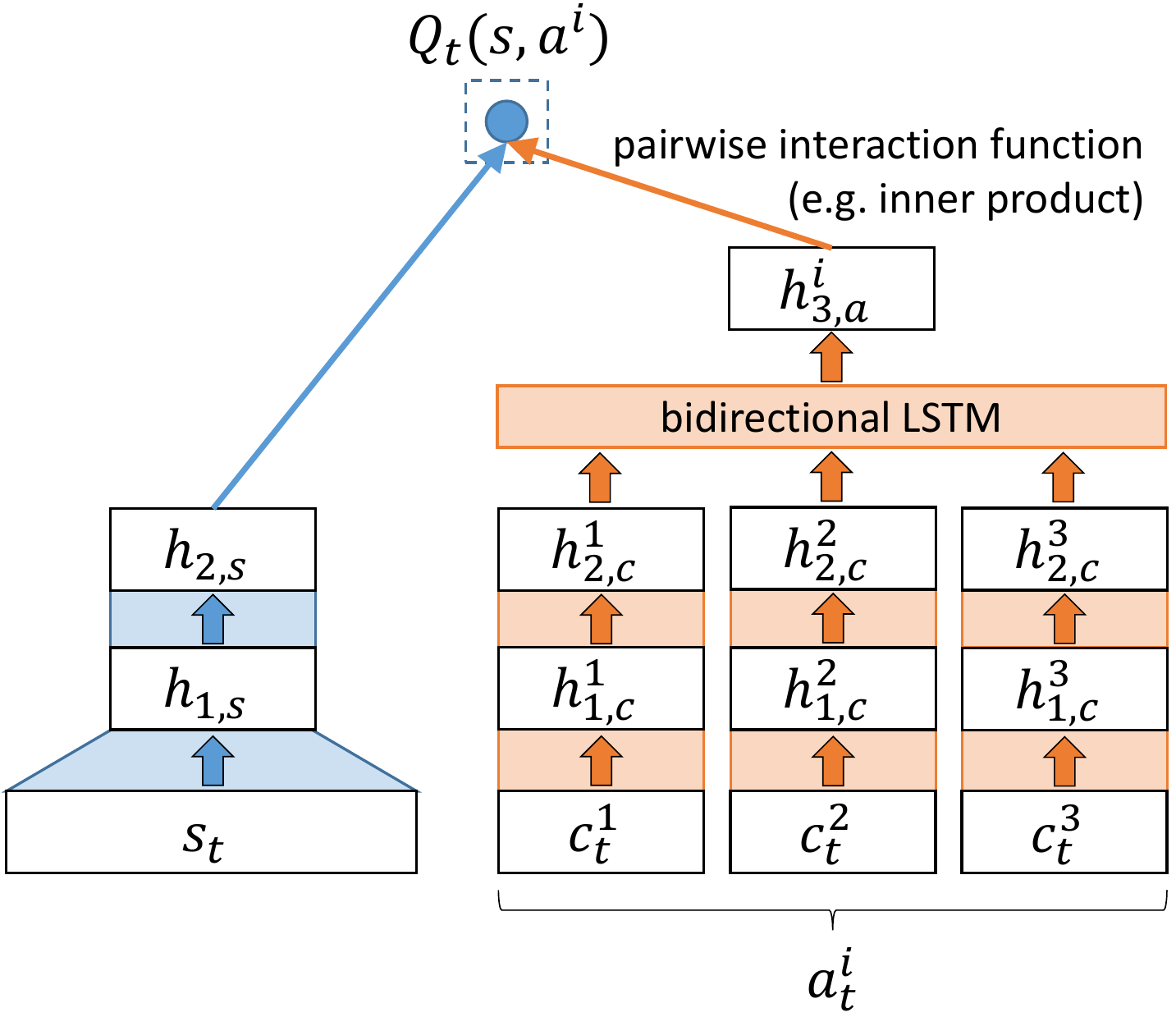}
	\label{fig:architectures_drrn-bilstm}
  	}
	\hfill
	\hfill
  }
\caption{Different deep Q-learning architectures}
\label{fig:architectures}
\end{figure*}


\section{Related Work}
\label{sec:related}
There has been increasing interest in applying deep reinforcement learning to a variety of problems, including tasks involving natural language. To control agents directly given high-dimensional sensory inputs, a Deep Q-Network \cite{mnih2015human} has been proposed and shown high capacity and scalability for handling a large state space. Another stream of work in recent deep learning research is the attention mechanism \cite{bahdanau2015neural,sukhbaatar2015end,vinyals2015grammar}, where a probability distribution is computed to pay attention to certain parts of a collection of data. It has been shown that the attention mechanism can handle long sequences or a large collection of data, while being quite interpretable. 
The attention mechanism work that is closest to ours is memory network (MemNN) \cite{weston2014memory,sukhbaatar2015end}. 
Most work on MemNNs uses embeddings of a query and documents to compute the attention weights for memory slots. Here, we propose models that also use non-content based features (time, popularity) for memory addressing. This helps retrieve content that provides complementary information to what is modeled in the query embedding vector. 
In addition, the content-based component of our query scheme uses TF-IDF based semantic-similarity, since the memory comprises a very large corpus of external documents that makes end-to-end learning of attention features impractical.

Multiple studies have explored interacting with a database (or knowledge base) using reinforcement learning. Narasimhan et al. \shortcite{narasimhan-yala-barzilay:2016:EMNLP2016} presents a framework of acquiring and incorporating external evidence to improve extraction accuracy in domains where the amount of training data is scarce. In task-oriented human-computer dialogue interactions, Wen et al. \shortcite{wen2016network} introduce a neural network-based trainable dialogue system with a database operator module. Dhingra et al. \shortcite{dhingra2016end} proposed a dialogue agent that provides users with an entity from a knowledge base by interactively asking for its attributes. In question answering, knowledge representation and reasoning also plays a central role \cite{ferrucci2010building,boyd2012besting}. Our goal differs from these studies in that we do not directly optimize a domain-specific knowledge search, instead we use external world knowledge to enrich the state representation in a reinforcement learning task.

Our task of tracking popular Reddit comments is somewhat related to an approach to multi-document summarization described in \cite{daume2009search}. A difference with respect to our problem is that the space of text for selection evolves over time.  In addition, in our case, the agent has no access to optimal policy, in contrast to the SEARN algorithm used in that work. 


To address overestimations of action values, double Q-learning \cite{Hasselt2010double,van2015deep} has been proposed and it leads to better performance gains on several Atari games. Dulac-Arnold et al. \shortcite{dulac2016reinforcement} present a policy architecture that works efficiently with a large number of actions. While a combinatorial action space can be large and discrete, this method does not directly apply in our case, because the possible actions are changing over different states. Instead, we borrow the philosophy from double Q and propose a two-stage Q-learning approach to reduce computational complexity by using a first Q-function to construct a quick yet rough estimate in the combinatorial action space, and then a second Q-function to rerank a set of sub-optimal actions.

The work described in our paper improves over \cite{he2016emnlp} by augmenting the state representation with external knowledge and by combining the two architectures that they proposed in two-stage Q-learning to enable exploration of the full action space. The first DRRN evaluates an action by treating sub-actions as independent and summing their contributions to the Q-value (Figure \ref{fig:architectures_drrn-sum}), and the second models potential redundancy of sub-actions by using a BiLSTM at the comment level (Figure \ref{fig:architectures_drrn-bilstm}).

\section{Incorporating External Knowledge into the State Representation}
\label{sec:external}
This approach is inspired by the observation that in a real-world decision making process, it is usually beneficial to consider background knowledge. 
Here, we introduce a mechanism to incorporate external language knowledge into decision making. The intuition is that the agent will keep track of a memory space that helps with decision making, and when a new state comes, the agent refers to this external knowledge and picks relevant resources to help with decision making.

\begin{figure}[t]
	\includegraphics[width=0.48\textwidth]{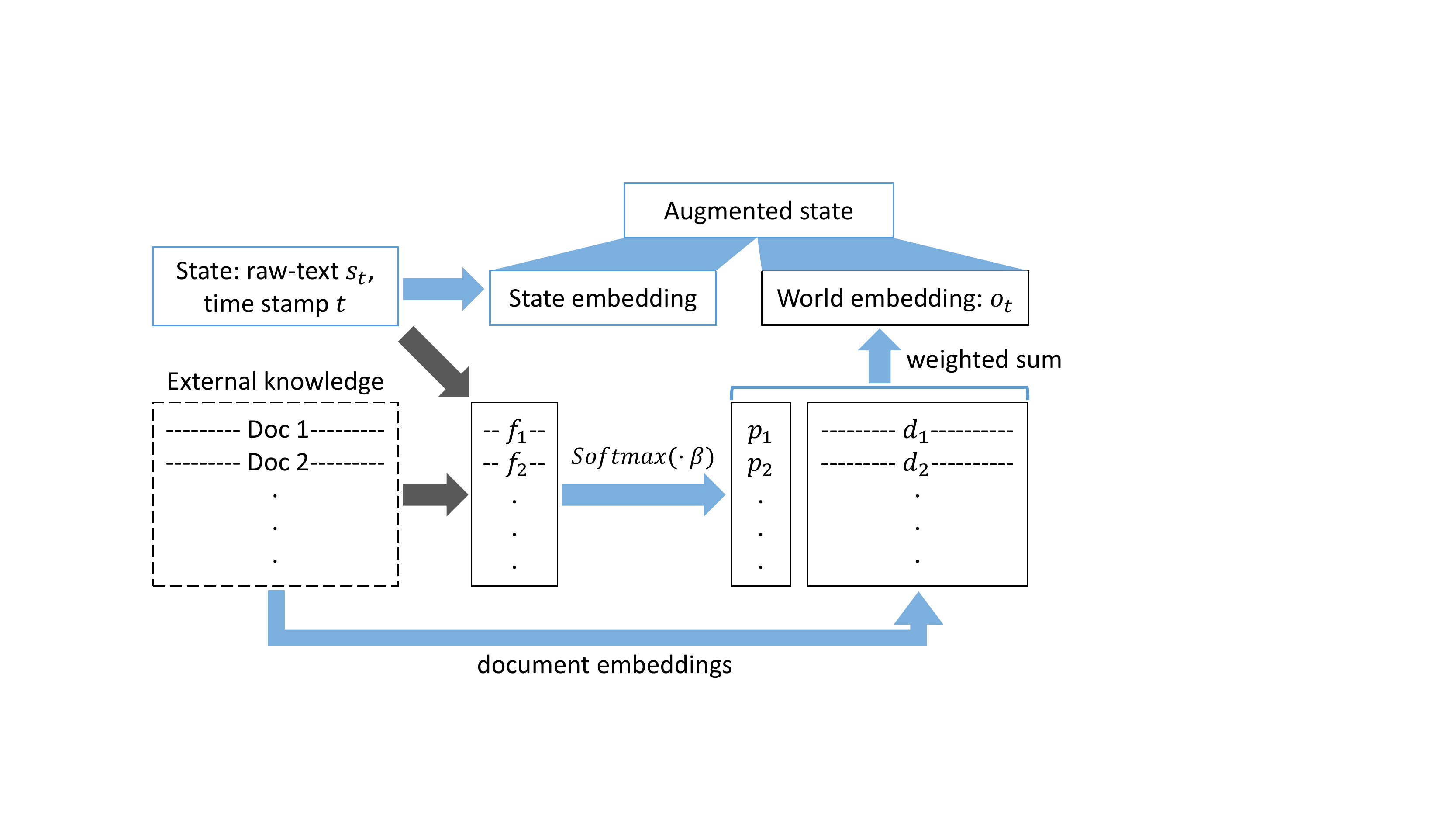}
\caption{Incorporating external knowledge to augment a state-side representation with an attention mechanism. The attention features $\{f_1, f_2, \cdots\}$ depend on the state and time stamp, helping the agent learn to pay different attention to external knowledge given different states. The shaded blue parts are learned end-to-end within reinforcement learning.}
\label{fig:DRRN.MemNN}
\end{figure}

The architecture we propose is illustrated in Figure \ref{fig:DRRN.MemNN}. Every time the agent reads the state information from the environment, it performs a lookup operation in external knowledge in its memory. This external knowledge could be a static knowledge base, or more generally it can be a dynamic database. In our experiments, the agent keeps an evolving collection of documents from the worldnews subreddit. We use an attention mechanism that produces a probability distribution over the entire external knowledge resource. This weight vector is computed by considering a set of features measuring the relevance between the current state and the ``world knowledge'' of the agent. More specifically, we consider the following three types of relevance:
\begin{itemize}
\item Timing features: when users express their opinions on a website such as Reddit, it is likely they are referring to more recent news events. We use two indicator features to represent whether a document from the external knowledge is within the past 24 hours, or the past 7 days relative to the time of the new state. We denote these features as $\mathbb{1}_\text{day}$ and $\mathbb{1}_\text{wk}$, respectively.
\item Semantic similarity: we use the standard tf-idf (term-frequency inverse-document-frequency) \cite{salton1986introduction} and compute cosine similarity scores as a measure for semantic relevance between the current state and each document in the external knowledge. We denote this semantic similarity as $u_\text{sem}\in [-1,1]$.
\item Popularity: for reddit posts/comments, we may use karma score as a measure for popularity. It is possible that high popularity topics will occur more often in the environment. To compensate the range difference in different relevance measures, we normalize karma scores\footnote{Detailed descriptions are given in Section \ref{sec:experiments}.} so the feature values fall in the range $[0,1]$. We denote this normalized popularity score as $u_\text{pop}$.
\end{itemize}

For each state the agent extracts the above features for each document in the external knowledge, and form a 4-dimensional feature vector $f=[\mathbb{1}_\text{day}, \mathbb{1}_\text{wk}, u_\text{sem}, u_\text{pop}]$. The attention weights are then computed as a linear combination followed by a softmax over the entire external knowledge:
\begin{align*}
\bm{p}=\text{Softmax}(\bm{[1_\text{day}, 1_\text{wk}, u_\text{sem}, u_\text{pop}] {\cdot} \beta})
\end{align*}
where the Softmax operates over the collection of documents and $\bm{p}$ has dimension equals the number of documents. Note in our experimental setting, the softmax applies for only documents that exist before the new comments appear, and this simulates a ``real-time'' dynamic external knowledge resource. The attention weights $\bm{p}$ are then multiplied with document embeddings $\{d_i\}$ to form a vector representation (embedding) of ``world'' knowledge:
\begin{align*}
o=\sum_i p_i d_i
\end{align*}
The world embedding is concatenated with the original state embedding to enrich understanding of the environment.


\section{Two-Stage Q-learning for a Combinatorial Action Space}
\label{sec:double_q}


There are two challenges associated with a combinatorial action space. One is the development of a Q-function framework for estimating the long-term reward. This is addressed in \cite{he2016emnlp}. The other is the potentially high computational complexity, due to evaluating $Q$ over every possible pair of $(s_t, a_t^i)$. In the case of deep Q-learning, most of the time has been spent on the forward-pass from $N\choose K$ actions to $N\choose K$ Q-values. For back-propagation, since we only need to back-propagate one particular action the agent has chosen, complexity is not affected by the combinatorial action space.

One solution to sidestep computational complexity is to randomly pick a fixed number, say $m$ candidate actions, and perform a $\max$ operation. While this is widely used in the reinforcement learning literature, it is problematic in our application because the large and highly skewed action space makes it likely that good actions are missed. Here we propose to use two-stage Q-learning for reducing search complexity. More specifically, we can rewrite the $\max$ operation as:
\begin{align*}
\max_{a_t\in \mathcal{A}_t} Q_2(s_{t}, a_t) \approx \max_{a_t \in \mathcal{B}_t} Q_2(s_{t}, a_t)
\end{align*}
where
\begin{align} \label{eq:1}
\mathcal{B}_t = \argmax_{a_t \in \mathcal{A}_t}^{m} Q_1(s_{t}, a_t)
\end{align}
where $\argmax_{a_t\in \mathcal{A}_t}^{m}$ means picking the top-$m$ actions from the whole action set $\mathcal{A}_t$.

In the case of $Q_1$ being DRRN-Sum, we can rewrite $Q_1(s_{t}, a_t)$ as:
\begin{align*}
Q_1(s_{t}, a_t) = \sum_{i=1}^K Q_0(s_{t}, c_t^i) = \sum_{i=1}^K q_{t}^i
\end{align*}
which is simplified by precomputing sub-action value $q_{t}^i = Q_0(s_{t}, c_{t}^i)$, $i=1, \cdots, N$. $Q_0$ is the simple DRRN introduced in He et al. \shortcite{he2016acl}.

To elaborate, the idea is to use a first Q function $Q_1$ to perform a quick but rough ranking of $a_t^i$. The second Q function $Q_2$, which can be more sophisticated, is used to rerank the top-$m$ candidate actions. This is effectively a beam search with coarse-to-fine models and reranking. This ensures that all comments are explored, and at the same time, the architecture can be sophisticated enough to capture detailed dependencies between sub-actions, such as information redundancy. In our experiments, we pick $Q_1$ to be DRRN-Sum and $Q_2$ to be DRRN-BiLSTM. While the independence assumption on sub-action interdependency is too strong, the DRRN-Sum model is relatively easy to train. Since the parameters on the action side are tied for different sub-actions, we can train a DRRN with $K=1$ and then apply the model for each pair of $(s_t, c_t^i)$. This will result in $N$ sub-action Q-values $Q_0(s_t, c_t^i), i=1, 2, \cdots, N$. Thus computing Equation \ref{eq:1} is equivalent to sorting $N\choose K$ values. Thus, we avoid the huge computational cost of first generating $N\choose K$ actions from $N$ sub-actions, then applying a general Q-function approximation to come up with $N\choose K$. In Section \ref{sec:experiments}, we train a DRRN (with $K=1$) and then copy the parameters to DRRN-Sum, which can be used to evaluate the full action space.\footnote{The whole two-stage Q framework is summarized in Algorithm 1 in Appendix.}


\section{Experiments}
\label{sec:experiments}
\subsection{Data set and preprocessing}
We carry out experiments on the task of predicting popular discussion threads on Reddit, as proposed by He et al. \shortcite{he2016emnlp}. Specifically, we conduct experiments on data from 5 subreddits including askscience, askmen, todayilearned, askwomen, and politics, which cover diverse genres and topics. In order to have long enough discussion threads, we filter out discussion trees with fewer than 100 comments. For each of the 5 subreddits, we randomly partition 90\% of the data for online training, and 10\% of the data for testing. Our evaluation metric is accumulated karma scores. For each setting we obtain mean (average reward) and standard deviation (shown as error bars or numbers in brackets) by 5 independent runs, each over 10,000 episodes. In all our experiments we set $N=10$. The basic subreddit statistics are shown in Table \ref{table:subreddit-stats}. We also report random policy performances and oracle upper bound performances.\footnote{Upper bounds are estimated by exhaustively searching through each discussion tree to find $K$ max karma discussion threads (overlapped comments are counted only once). This upper bound may not be attainable in a real-time setting. For askscience, $N=10$ and different $K$'s, the upper bound performances range from 1991.3 ($K=2$) to 2298.0 ($K=5$).}

\begin{table}
\small
\centering
\begin{tabular}{| p{1.6cm} | P{0.95cm} | P{1.0cm} | P{1.1cm} | P{1.1cm} |} \hline
\bf Subreddit & \bf \# Posts (in k) & \bf \# Comments (in M) & \bf Random & \bf Upper bound \\ \hline
askscience & 0.94 & 0.32 & 321.3 & 2109.0 \\ \hline
askmen & 4.45 & 1.06 & 132.4 & 651.4 \\ \hline
todayilearned & 9.44 & 5.11 & 390.3 & 2679.6 \\ \hline
askwomen & 3.57 & 0.81 & 132.4 & 651.4 \\ \hline
politics & 4.86 & 2.18 & 149.3 & 967.7 \\ \hline \hline
worldnews & 9.88 & 5.99 & 205.8 & 1853.4 \\ \hline
\end{tabular}
\caption{Basic statistics of filtered subreddits}
\label{table:subreddit-stats}
\end{table}



In preprocessing we remove punctuation and lowercase all words. We use a bag-of-words representations for each state $s_t$, and comment $c_t^i$ in discussion tracking, and for each document in the external knowledge source. The vocabulary contains the most frequent 5,000 words and the out-of-vocabulary rate is 7.1\%. We use fully-connected feed-forward neural networks to compute state, action and document embeddings, with $L=2$ hidden layers and hidden dimension 20.

Our Q-learning agent uses $\epsilon$-greedy ($\epsilon=0.1$) throughout online training and testing. The discounting factor $\gamma=0.9$. During training, we use experience replay \cite{lin1992self} and the memory size is set to 10,000. For each experience replay, 500 episodes are generated and tuples are stored in a first-in-first-out fashion. We use mini-batch stochastic gradient descent with batch size of 100, and constant learning rate $\eta=0.000001$. We train separate models for different subreddits.

\subsection{Incorporating external knowledge}
We first study the effect of incorporating external knowledge, without considering the combinatorial action space. More specifically, we set $K=1$ and use the simple DRRN. Each action is to pick a comment $\{c_t^1\}$ from $\mathcal{C}_t$ to track. Our proposed method uses a state representation augmented by the world knowledge, as illustrated in Figure \ref{fig:DRRN.MemNN}.

We utilize the worldnews subreddit as our external knowledge source. This subreddit consists of 9.88k posts. We define each document in the world knowledge to be the post plus its top-5 comments ranked by karma scores. The agent keeps a growing collection of documents. That is, at each time $t$, the external knowledge contains documents from worldnews that appear before time $t$. To compute popularity score of each document, we simply sum the karma scores of post and top-5 comments. Then the karma scores are normalized by dividing the highest score in the external knowledge.\footnote{Unlike in Fang et al. \shortcite{ostendorf2016learning}, the summed karma scores do not follow a Zipfian distribution, so we do not use quantization or any nonlinear transformation.} Thus the popularity feature values for computing attention fall in the range $[0, 1]$.

\begin{figure*}[t]
	\centering
	\includegraphics[width=0.9\textwidth]{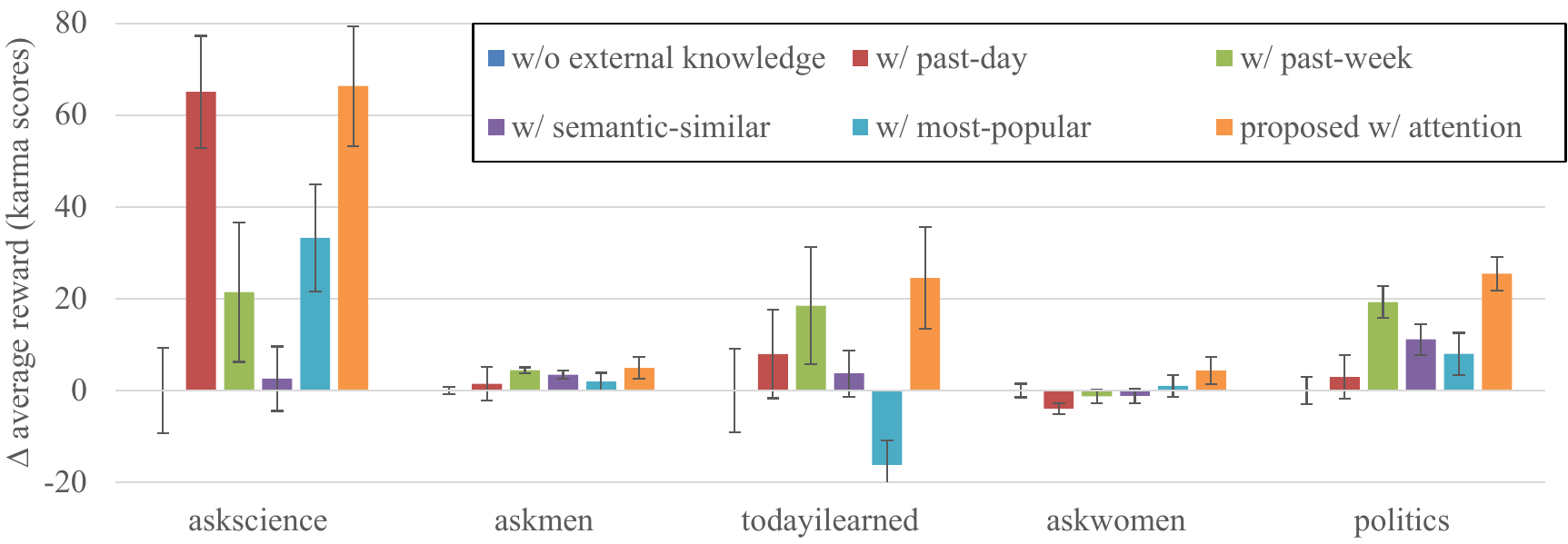}
\caption{DRRN (with multiple ways of incorporating external knowledge) performance gains over baseline DRRN (without external knowledge) across 5 different subreddits}
\label{fig:DRRN.external.results}
\end{figure*}

For comparison, we experiment with a baseline DRRN without any external knowledge. We also construct a baseline DRRN with hand-crafted rules for picking documents from external knowledge. Those rules include: i) documents within the past-day, ii) documents within the past-week, iii) 10 semantically most similar documents, iv) 10 most popular documents. We use a bag-of-words representation and construct the world embedding used to augment the state representation.

We compare multiple ways of incorporating external knowledge for different subreddits and show performance gains over a baseline DRRN (without any external knowledge) in Figure \ref{fig:DRRN.external.results}. The experimental results show that the DRRN using a learned attention mechanism to retrieve relevant knowledge outperforms all other configurations of DRRNs with rules for knowledge retrieval, and significantly outperforms the DRRN baseline that does not use external knowledge. Also we observe that different relevance features have different impact across subreddits. For example, for askscience, past-day documents have higher impact than past-week documents, while for politics past-week documents are more important. The most-popular documents actually have a negative effect for todayilearned, mainly because those are documents which are most popular throughout the entire history, while todayilearned discussions value information about recent events.\footnote{In principle, since we are concatenating the world embedding to obtain an augmented state representation, the result should not get worse. We hypothesize this is due to overfitting and use of mismatched documents, as in the most-popular setting for todayilearned.} Nevertheless, the attention mechanism learns to rely on proper features to retrieve useful knowledge for the needs of different domains.


\subsection{Two-stage Q-learning for a combinatorial action space}
In this subsection we study the effect of two-stage Q-learning, without considering external knowledge. 
We train DRRN ($K=1$) first, and copy over the parameters to DRRN-Sum as $Q_1$. We then train $Q_2$=DRRN-BiLSTM as before, except that we use $Q_1$=DRRN-Sum to explore the whole action space to obtain $\mathcal{B}_t$.

\begin{table}
\small
\centering
\begin{tabular}{| p{0.4cm} | c | c | c |} \hline
$\mathcal{B}_t$ & random & all & DRRN-Sum \\ \hline
$\bm Q_2$ & DRRN-BiLSTM & DRRN-Sum & DRRN-BiLSTM \\ \hline \hline
K=2 & 573.2 (12.9) & 663.3 (8.7) & \bf 676.9 (5.5) \\ \hline
K=3 & 711.1 (8.7) & 793.1 (8.1) & \bf 833.9 (5.7) \\ \hline
K=4 & 854.7 (16.0) & 964.5 (12.0) & \bf 987.1 (12.1) \\ \hline
K=5 & 980.9 (21.1) & 1099.4 (15.9) & \bf 1101.3 (13.8) \\ \hline
\end{tabular}
\caption{A performance comparison (across different $K$'s on askscience subreddit)}
\label{table:askscience-double-q}
\end{table}

On askscience, we try multiple settings with $K=2, 3, 4, 5$ and the results are shown in Table \ref{table:askscience-double-q}. We compare the proposed two-stage Q-learning with two single-stage Q-learning baselines. The first baseline, following the method in He et al. \shortcite{he2016emnlp}, uses a random subsampling approach to obtain $\mathcal{B}_t$ (with $m=10$) and takes the max over them using DRRN-BiLSTM. The second baseline uses DRRN-Sum and explores the whole action space.
The proposed two-stage Q-learning uses DRRN-Sum for picking a $\mathcal{B}_t$ and DRRN-BiLSTM for reranking. We observe a large improvement by switching from ``random'' to ``all'', showing that exploring the entire action space is critical in this task. There is a consistent gain by using two-stage Q-learning instead of a single-stage Q with DRRN-Sum. This shows that using a more sophisticated value function for reranking also helps with performance.


\begin{table*}
\small
\centering
\begin{tabular}{| l | l || >{\centering}p{1.8cm} | >{\centering}p{1.8cm} | c | c | c |} \hline
$\mathcal{B}_t$ & $\bm Q_2$ & askscience & askmen & todayilearned & \ askwomen\  & \ \ \ politics\ \ \  \\ \hline
random & DRRN-BiLSTM & 711.1 (8.7) & 139.0 (3.6) & 606.9 (15.8) & 135.0 (1.3) & 177.9 (3.3) \\ \hline
all & DRRN-Sum & 793.1 (8.1) & 142.5 (2.3) & 679.4 (11.4) & 145.9 (2.4) & 180.6 (6.3) \\ \hline
DRRN-Sum & DRRN-BiLSTM & \bf 833.9 (5.7) & \bf 148.0 (5.5) & \bf 697.9 (9.4) & \bf 149.6 (3.3) & \bf 204.7 (4.2) \\ \hline
\end{tabular}
\caption{A performance comparison (across different subreddits) with $K=3$}
\label{table:subreddit-double-q}
\end{table*}

In Table \ref{table:subreddit-double-q}, we compare two-stage Q-learning with the two baselines across different subreddits, with $N=10, K=3$. The findings are consistent with those for askscience. Since different subreddits may have very different karma score distributions and language style, our results suggest that the algorithm applies well to different community interaction styles. 

During testing, we compare runtime of the DRRN-BiLSTM Q-function with different $\mathcal{B}_t$, simulating over 10,000 episodes with $N=10$ and $K=2,3,4,5$. The search time for the random selection and the two-stage Q-function are similar, both nearly constant for different $K$. Using two-stage Q the test runtime is reduced by $6\times$ for $K=3$ and $11\times$ for $K=5$ comparing to exploring the whole action space.\footnote{Training DRRN-BiLSTM with the whole action space is intractable, so we just used a subspace trained DRRN-BiLSTM model for testing. This however achieves worse performance compared to the two-stage Q probably due to mismatch in training and testing.}



\subsection{Combined results}

\begin{figure}[t]
  \centering
  \includegraphics[width=0.5\textwidth]{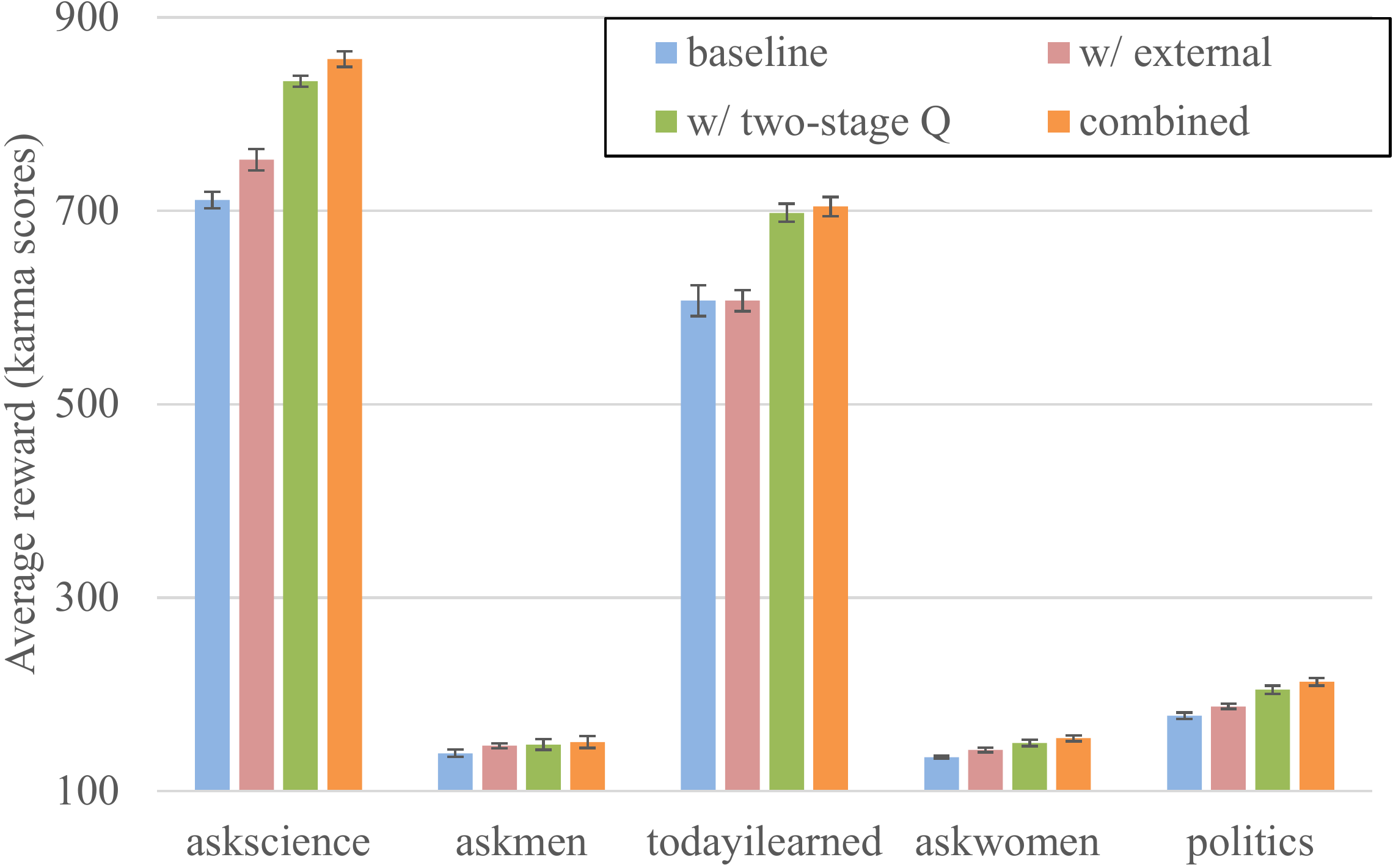}
\caption{Ablation study on effects by incorporating external knowledge and/or two-stage Q-learning across 5 different subreddits.}
\label{fig:final.combined.results}
\end{figure}

\begin{table*}
\small
\centering
\begin{tabular}{| p{1.4cm} || p{3.2cm} | p{3.2cm} | p{3.4cm} | p{2.5cm} |} \hline
state & top-1 & top-2 & top-3 & least \\ \hline
Would it be possible to artificially create an atmosphere like Earth has on Mars? & Ultimate Reality TV: A Crazy Plan for a Mars Colony - It might become the mother of all reality shows. Fully 704 candidates are soon to begin competing for a trip to Mars to establish a colony there. & `Alien thigh bone' on Mars: Excitement from alien hunters at `evidence' of extraterrestrial life. Mars likely never had enough oxygen in its atmosphere and elsewhere to support more complex organisms. & The Gaia (General Authority on Islamic Affairs) and the UAE (United Arab Emirates) have issued a fatwa on people living on mars, due to the religious reasoning that there is no reason to be there. & North Korea's internet is offline; massive DDOS attack presumed. \\ \hline
Does our sun have any unique features compared to any other star? & Star Wars: Episode VII begins filming in UAE desert.      This can't possibly be a modern Star Wars movie! I don't see a green screen in sight!   Ya, it's more like Galaxy news. & African Pop Star turns white (and causes controversy) with new line of skin whitening cream. I would like to see an unshopped photo of her in natural lighting. & Dwarf planet discovery hints at a hidden Super Earth in solar system - The body, which orbits the sun at a greater distance than any other known object, may be shepherded by an unseen planet. & Hong Kong democracy movement hit by 2018.  The vote has no standing in law, by attempting to sabotage it, the Chinese(?) are giving it legitimacy   \\ \hline
\end{tabular}
\caption{States and documents (partial text) showing how the agent learns to attend to different parts of external knowledge}
\label{table:attention.documents}
\end{table*}

In Figure \ref{fig:final.combined.results}, we present an ablation study on effects of incorporating external knowledge and/or two-stage Q-learning (with $N=10, K=3$) across different subreddits. The two contributions we proposed each help improve reinforcement learning performance in a natural language scenario with a  combinatorial action space. In addition, combining these two approaches further improves performance. In our task, two-stage Q-learning provides a larger gain. However, in all cases, incorporating external knowledge consistently gives additional gain on top of two-stage Q-learning.

We conduct case studies in Table \ref{table:attention.documents}. We show examples of most/least attended documents in the external knowledge given the state description. The documents are shortened for brevity. In the first example, the state is about a question about the atmosphere on Mars. The most-attended documents are correctly related to Mars living conditions, in various sources and aspects. The second example has the state talking about sun's features compared to other stars. Interestingly, although the agent is able to attend to top documents due to some topic word matching (e.g. sun, star), the picked documents reflect popularity more than topic relevance. 
The least-attended documents are totally irrelevant in both examples, as expected.


\section{Conclusion}
\label{sec:conclusion}
In this paper we introduce two approaches for improving natural language based decision making in a combinatorial action space. The first is to augment the state representation of the environment by incorporating external knowledge through a learnable attention mechanism. The second is to use a two-stage Q-learning framework for exploring the entire combinatorial action space, while avoiding enumeration of all possible action combinations. Our experimental results show that both proposed approaches improve the performance in the task of predicting popular Reddit threads.

\bibliography{emnlp2017,refs}
\bibliographystyle{emnlp_natbib}

\newpage
\section*{\Large Supplementary Material}
\appendix

\section{Algorithm table for two-stage Q-learning}
As shown in Algorithm \ref{Alg:double-q-learning}.

\begin{algorithm*}[b!]
\begin{algorithmic}[1]
\STATE Initialize Reddit popularity prediction environment and load dictionary.
\STATE Initialize DRRN $Q_0(s_t, c_t^i; \Theta_1)$ (equivalent as DRRN-Sum with $K=1$) with small random weights and train. The DRRN-Sum $Q_1(s_t, a_t; \Theta_1)=Q_1(s_t, \{c_t^1, c_t^2, \cdots, c_t^K\}; \Theta_1)=\sum_{i=1}^K {Q_0(s_t, c_t^i; \Theta_1)}$ shares the same parameters as DRRN.
\STATE Initialize replay memory $\mathcal{D}$ to capacity $|\mathcal{D}|$.
\FOR {$episode = 1, \ldots, M$}
	\STATE Randomly pick a discussion tree.
	\STATE Read raw state text and a list of sub-action text from the simulator, and convert them to representation $s_1$ and $c_{1,1}, c_{1,2}, \ldots, c_{1,N}$.
	\STATE Compute $q_{1,j}=Q_0(s_1, c_{1,j}; \Theta_1)$ for the list of sub-actions using DRRN forward activation.
	\STATE For each $a_1\in \mathcal{A}_1$, form value of $Q_1(s_1, a_1; \Theta_1)=\sum_{i=1}^K {Q_0(s_1, c_1^i; \Theta_1)}=\sum_{i=1}^K q_1^i$.
	\STATE Keep a list of top $m$ actions $\mathcal{B}_1=[a_1^1, a_1^2, \cdots, a_1^{m}]$, where each $a_1^i$ consists of $K$ sub-actions.
	\FOR{$t = 1, \ldots, T$}
		\STATE Compute $Q_2(s_t, a_t^i; \Theta_2), i=1, 2, \cdots, m$ for $\mathcal{B}_t$, the list of top $m$ actions using DRRN-BiLSTM forward activation.
		\STATE Select an action $a_t$ based on policy $\pi(a_t = a_t^i | s_t)$ derived from $Q_2$. Execute $a_t$ in simulator.
		\STATE Observe reward $r_{t+1}$. Read the next state text and the next list of sub-action texts, and convert them to representation $s_{t+1}$ and $c_{t+1,1}, c_{t+1,2}, \ldots, c_{t+1,N}$.
		\STATE Compute $q_{t+1,j}=Q_0(s_{t+1}, c_{t+1,j}; \Theta_1)$ for the list of sub-actions using DRRN.
		\STATE For each $a_{t+1}\in \mathcal{A}_{t+1}$, form value of $Q_{t+1}(s_{t+1}, a_{t+1}; \Theta_1)=\sum_{i=1}^K {Q_0(s_{t+1}, c_{t+1}^i; \Theta_1)}=\sum_{i=1}^K q_{t+1}^i$.
		\STATE Keep a list of top $m$ actions $\mathcal{B}_{t+1}=[a_{t+1}^1, a_{t+1}^2, \cdots, a_{t+1}^{m}]$, where each $a_{t+1}^i$ consists of $K$ sub-actions.
		\STATE Store transition $(s_t, a_t, r_{t+1}, s_{t+1}, \mathcal{B}_{t+1})$ in $\mathcal{D}$.
		\IF {during training}
			\STATE Sample random mini batch of transitions $(s_k, a_k, r_{k+1}, s_{k+1}, \mathcal{B}_{k+1})$ from $\mathcal{D}$.
			\STATE Set $y_k=\begin{cases}
					r_{k+1} & \text{if } s_{k+1} \text{ is terminal}\\
					r_{k+1}+\gamma\max_{a'\in \mathcal{B}_{k+1}}Q_2(s_{k+1}, a'; \Theta_2)) & \text{otherwise}
				\end{cases}$
			\STATE Perform a gradient descent step on $(y_k-Q_2(s_k, a_k; \Theta_2))^2$ with respect to the network parameters $\Theta_2$. Back-propagation is performed only for $a_k$ though there are $|\mathcal{A}_{k}|$ actions.
		\ENDIF
	\ENDFOR
\ENDFOR
\end{algorithmic}
\caption{Two-stage Q-learning in combinatorial action space ($Q_1$: DRRN-Sum, $Q_2$: DRRN-BiLSTM)}
\label{Alg:double-q-learning}
\end{algorithm*}

\section{URLs for subreddits used in this paper}
As shown in Table \ref{table:subreddit-urls}. All post ids will be released for future work on this task.

\begin{table}[h!]
\small
\centering
\begin{tabular}{| l | l |} \hline
\bf Subreddit & \bf URL \\ \hline
askscience & https://www.reddit.com/r/askscience/ \\ \hline
askmen & https://www.reddit.com/r/askmen/ \\ \hline
todayilearned & https://www.reddit.com/r/todayilearned/ \\ \hline
askwomen & https://www.reddit.com/r/askwomen/ \\ \hline
politics & https://www.reddit.com/r/politics/ \\ \hline
worldnews & https://www.reddit.com/r/worldnews/ \\ \hline
\end{tabular}
\caption{URLs of subreddit data sets}
\label{table:subreddit-urls}
\end{table}

\end{document}